\documentclass[10pt, conference, compsocconf]{IEEEtran}
\IEEEoverridecommandlockouts
\usepackage{cite}
\usepackage{amsmath,amssymb,amsfonts}
\usepackage{algorithmic}
\usepackage{graphicx}
\usepackage{textcomp}
\usepackage{xcolor}
\usepackage{microtype}
\usepackage{graphicx}
\usepackage{booktabs}
\usepackage{subcaption}
\usepackage[bookmarks=false]{hyperref}
\usepackage{filecontents}
\usepackage{multirow}

\begin{document}

\title{Universal Barcode Detector via Semantic Segmentation}


\author{\IEEEauthorblockN{Andrey Zharkov\IEEEauthorrefmark{1}\IEEEauthorrefmark{2}, 
Ivan Zagaynov\IEEEauthorrefmark{1}\IEEEauthorrefmark{2}}
\IEEEauthorblockA{\IEEEauthorrefmark{1}
R\&D Department\\
ABBYY Production LLC\\
Moscow, Russia\\
\{andrew.zharkov, Ivan.Zagaynov\}@abbyy.com}
\IEEEauthorblockA{\IEEEauthorrefmark{2}
Phystech School of Applied Mathematics and Informatics\\
Moscow Institute of Physics and Technology (National Research University)
}}

\maketitle

\begin{abstract}
Barcodes are used in many commercial applications, thus fast and robust reading is important. There are many different types of barcodes, some of them look similar while others are completely different. In this paper we introduce new fast and robust deep learning detector based on semantic segmentation approach. It is capable of detecting barcodes of any type simultaneously both in the document scans and in the wild by means of a single model. The detector achieves state-of-the-art results on the ArTe-Lab 1D Medium Barcode Dataset with detection rate 0.995. Moreover, developed detector can deal with more complicated object shapes like very long but narrow or very small barcodes. The proposed approach can also identify types of detected barcodes and performs at real-time speed on CPU environment being much faster than previous state-of-the-art approaches.
\end{abstract}

\section{Introduction}
\begin{figure}
    \centering
    \begin{subfigure}[b]{0.33\linewidth}
        \includegraphics[width=0.9\textwidth]{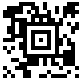}
        \caption{Aztec}
        \label{fig:gull}
    \end{subfigure}%
    \begin{subfigure}[b]{0.33\linewidth}
        \includegraphics[width=0.9\textwidth]{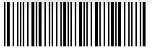}
        \caption{Codabar}
        \label{fig:gull2}
    \end{subfigure}%
    \begin{subfigure}[b]{0.33\linewidth}
        \includegraphics[width=0.9\textwidth]{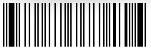}
        \caption{Code 93}
        \label{fig:gull2}
    \end{subfigure}%
    
    \begin{subfigure}[b]{0.33\linewidth}
        \includegraphics[width=0.9\textwidth]{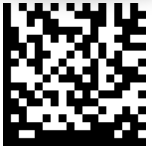}
        \caption{DataMatrix}
        \label{fig:gull2}
    \end{subfigure}%
    \begin{subfigure}[b]{0.33\linewidth}
        \includegraphics[width=0.9\textwidth]{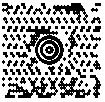}
        \caption{MaxiCode}
        \label{fig:gull2}
    \end{subfigure}%
    \begin{subfigure}[b]{0.33\linewidth}
        \includegraphics[width=0.9\textwidth]{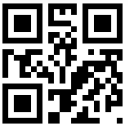}
        \caption{QRCode}
        \label{fig:gull2}
    \end{subfigure}%
    
    \begin{subfigure}[b]{0.33\linewidth}
        \includegraphics[width=0.9\textwidth]{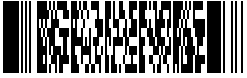}
        \caption{PDF417}
        \label{fig:gull2}
    \end{subfigure}%
    \begin{subfigure}[b]{0.33\linewidth}
        \includegraphics[width=0.9\textwidth]{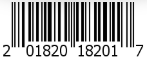}
        \caption{UPC-A}
        \label{fig:gull2}
    \end{subfigure}%
    \begin{subfigure}[b]{0.33\linewidth}
        \includegraphics[width=0.9\textwidth]{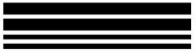}
        \caption{PatchCode}
        \label{fig:gull2}
    \end{subfigure}%
    
    \begin{subfigure}[b]{0.5\linewidth}
        \includegraphics[width=0.9\textwidth]{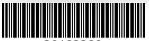}
        \caption{Standard 2 of 5}
        \label{fig:gull2}
    \end{subfigure}%
    \begin{subfigure}[b]{0.5\linewidth}
        \includegraphics[width=0.9\textwidth]{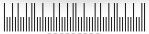}
        \caption{Postnet}
        \label{fig:gull2}
    \end{subfigure}%
    \caption{Some examples of different barcodes}
    \label{fig:barcode_samples}
\end{figure}

Starting from the 1960s people have invented many barcode types which serve for machine readable data representation and have lots of applications in various fields. The most frequently used are probably UPC and EAN barcodes for consumer products labeling, EAN128 serves for transferring information about cargo between enterprises, QR codes are widely used to provide links, PDF417 has variety of applications in transport, identification cards and inventory management. Barcodes have become ubiquitous in modern world, they are used as electronic tickets, in official documents, in advertisement, healthcare, for tracking objects and people. Examples of popular barcode types are shown in Fig.~\ref*{fig:barcode_samples}.

There are two main approaches for decoding barcodes, the  former uses laser  and the latter just a simple  camera. Through years of development, laser scanners have become very reliable and fast for the case of reading exactly one 1D barcode, but they are completely unable to deal with 2D barcodes or read several barcodes at the same time. Another drawback is that they can not read barcodes from screens efficiently as they strongly rely on reflected light. 

Popular camera-based reader is a simple smartphone application which is capable of scanning almost any type of barcode. However, most applications require some user guidance like pointing on barcode to decode. Most applications decode only one barcode at a time, despite it is possible to decode all barcodes in the image. It may become important when we need to scan barcodes from some official documents where might be a number of them.

In this work, we introduce segmentation based barcode detector which is capable of locating all barcodes simultaneously no matter how many of them are present in the image or which types they are, so the system does not need any user guidance. The developed detector also provides information about most probable types of detected barcodes thus decreasing time for reading process.

\section{Related Work}

The early work in the domain of barcode detection from 2D images was motivated by the wide spread of mobile phones with cameras. \cite{Ohbuchi} proposes a method for finding 2D barcodes via corner detection and 1D barcodes through spiral scanning. \cite{Wachefeld} introduces another method for 1D barcode detection based on decoding. Both approaches however require certain guidance from the user.

In more recent papers authors pay more attention on developing solutions which can be done automatically with less user guidance. \cite{Gallo2011Reading1B} finds regions with high difference between x and y derivatives, \cite{Tekin2013} calculates oriented histograms to find patches with dominant direction, \cite{Katona2013Efficient1A} relies on morphology operations to detect both 1D and 2D barcodes, reporting high accuracy on their own data. The work of S\"{o}r\"{o}s \emph{et al.} \cite{Soros2013} is notable as they compare their own algorithm with other works mentioned in this paragraph. They demonstrate that their approach is superior on the same dataset WWU Muenster Barcode Database (Muenster). Their algorithm is based on the idea that 1D barcodes have many edges, 2D barcodes have many corners, while text areas have both many edges and many corners.

The work of Cresot \emph{et al.}, 2015 \cite{Cresot2015} is a solid baseline for 1D barcode detection. They evaluated their approach on Muenster and on extended ArTe-Lab 1D Medium barcode database (Artelab) provided by Zamberletti \emph{et al.} \cite{Zamberletti_2013} outperforming him on both datasets. The solution in \cite{Cresot2015} seems to outperform \cite{Soros2013} despite it is hard to compare as they were evaluated on different datasets using slightly different metrics. Cresot's algorithm detects dark bars of barcodes using Maximal Stable Extremal Regions (MSER) followed by finding imaginary perpendicular to bars center line in Hough space. In 2016 Cresot \emph{et al.} came with a new paper \cite{Cresot2016} improving previous results using a new variant of Line Segment Detector instead of MSER, which they called Parallel Segment Detector. \cite{Namane2017} proposes another bars detection method for 1D barcode detection, which is reported to be absolutely precise in real-time applications.

In the recent years neural networks show very promising results in many domains including Computer Vision. However, at the moment of writing there are only a few research works which utilize deep learning for barcode detection. The first is already mentioned \cite{Zamberletti_2013} where neural network analyzes Hough space to find potential bars. More recent and promising results are obtained in \cite{YoloBarcode} where authors use YOLO (You Only Look Once) detector to find rectangles with barcodes, then apply another neural network to find the rotation angle of that barcode, thus after that they are able to rotate the barcode and pass it to any barcode recognizer simplifying the recognition task. They showed new state-of-the-art (SOTA) results on Muenster dataset. However, their solution can not be considered real-time for CPUs. Moreover, YOLO is known to have problems with narrow but very long objects, which can be an issue for some barcode types.

\section{Detection via segmentation}

Our approach is inspired by the idea of PixelLink \cite{Deng2018PixelLinkDS} where authors solve text detection via instance segmentation. We believe that for barcodes the situation when 2 of them are close to each other is unusual, so we do not really need to solve instance segmentation problem therefore dealing with semantic segmentation challenge should be enough. 

PixelLink shows good results capturing long but narrow lines with text, which can be a case for some barcode types so we believe such object shape invariance property is an additional advantage.

To solve the detection task we first run semantic segmentation network and then postprocess its results.

\subsection{Semantic segmentation network}

\begin{table*}[t]
\caption{Model architecture, C=24 is the number of channels and N is the number of predicted classes (barcode types)}
\label{segmentation_network}
\begin{center}
\begin{small}
\begin{sc}
\begin{tabular}{l|ccc|cccccc|c}
\toprule
& \multicolumn{3}{c|}{Downscale Module} & \multicolumn{6}{c|}{Context Module} & Final \\
Layer & 1 & 2 & 3 & 4 & 5 & 6 & 7 & 8 & 9 & 10 \\
\midrule
Stride & 2 & 1 & 2 & 1 & 1 & 1 & 1 & 1 & 1 & 1 \\
Dilation & 1 & 1 & 1 & 1 & 2 & 4 & 8 & 16 & 1 & 1 \\
Separable & Yes & Yes & Yes & No & No & No & No & No & No & No \\
Kernel & 3x3 & 3x3 & 3x3 & 3x3 & 3x3 & 3x3 & 3x3 & 3x3 & 3x3 & 1x1 \\
Output Channels & C & C & C & C & C & C & C & C & C & 1+N \\
Receptive Field & 3x3 & 7x7 & 11x11 & 19x19 & 35x35 & 67x67 & 131x131 & 259x259 & 267x267 & 267x267 \\
\bottomrule
\end{tabular}
\end{sc}
\end{small}
\end{center}
\vskip -0.15in
\end{table*}

Barcodes normally can not be too small so predicting results for resolution 4 times lower than original image should be enough for reasonably good results. Thus we find segmentation map for superpixels which are 4x4 pixel blocks.

Detection is a primary task we are focusing on in this work, treating type classification as a less important sidetask. Most of barcodes share a common structure so it is only natural to classify pixels as being part of barcode (class 1) or background (class 0), thus segmentation network solves binary (super)pixel classification task.

Barcodes are relatively simple objects and thus may be detected by relatively simple architecture. To achieve real-time CPU speed we have developed quite simple architecture based on dilated and separable convolutions (see \autoref*{segmentation_network}). It can be logically divided into 3 blocks:
\begin{enumerate}
    \item \textbf{Downscale Module} is aimed to reduce spatial features dimension. Since these initial convolutions are applied to large feature maps they cost significant amount of overall network time, so to speed up inference these convolutions are made separable.
    
    \item \textbf{Context Module}. This block is inspired by \cite{Yu2015MultiScaleCA}. However, in our architecture it serves for slightly different purpose just improving features and exponentially increasing  receptive  field  with  each  layer.
    
    \item Final classification layer is 1x1 convolution with number of filters equal to 1+n\_classes, where n\_classes is number of different barcode types we want to differentiate with.
\end{enumerate}

We used ReLU nonlinearity after each convolution except for the final one where we apply sigmoid to the first channel and softmax to all of the rest channels.

We have chosen the number of channels $C=24$ for all convolutional layers. Our experiments show that with more filters model has comparable performance, but with less filters performance drops rapidly. As we have only a few channels in each layer the final model is very compact with only 32962 weights.

As the maximal image resolution we are working with is 512x512, receptive field for prediction is at least half an image which should be more than enough contextual information for detecting barcodes.

\subsection{Detecting barcodes based on segmentation}

After the network pass we get segmentation map for superpixels with $1+n\_classes$ channels. For detection we use only the first channel which can be interpreted as probability being part of barcode for superpixels.

We apply the threshold value for probability to get detection class binary labels (barcode/background). In all our experiments we set this threshold value to 0.5. We now find connected components on received superpixel binary mask and calculate bounding rectangle of minimal area for each component. To do the latest we apply \textit{minAreaRect} method from \textit{OpenCV} library (accessed Dec 2018).

Now we treat found rectangles as detected barcodes. To get detection rectangle on original image resolution we multiply all of its vertices coordinates by the network scale 4.

\subsection{Filtering results}

To avoid a situation when a small group of pixels is accidentally predicted as a barcode, we filter out all superpixel connected components with area less than threshold $T_{area}$. The threshold value should be chosen to be slightly less than minimal area of objects in the dataset on the segmentation map. In all of our experiments we used value $T_{area}=20$.

\subsection{Classification of detected objects}

To determine barcode type of detected objects we use all of the rest $n\_classes$ channels from segmentation network output. After softmax we treat them as probabilities of being some class.

Once we found the rectangle we compute the average probability vector inside this rectangle, then naturally choose the class with the highest probability.

\section{Optimization scheme}

\subsection{Loss function}

The training loss is a weighted sum of detection and classification losses

\begin{equation}
    L = L_{detection} + \alpha L_{classification}
\end{equation}

Detection loss $L_{detection}$ itself is a weighted sum of three components: mean binary crossentropy loss on positive pixels $L_{p}$, mean binary crossentropy loss on negative pixels $L_{n}$, and mean binary crossentropy loss on worst predicted $k$ negative pixels $L_{h}$, where $k$ is equal to the number of positive pixels in image.

\begin{equation}
    L_{detection} = w_{p} L_{p} + w_{n} L_{n} + w_{h} L_{h}
\end{equation}

Classification loss is mean (categorical) crossentropy computed by all channels except the first one (with detection). Classification loss is calculated only on superpixels which are parts of ground truth objects.

As our primary goal is high recall in detection we have chosen $w_p=15$, $w_n=1$, $w_h=5$, $\alpha=1$. We also tried several different configurations but this combination was the best among them. However, we did not spent too much time on hyperparameter search.

\subsection{Data augmentation}

For augmentation we do the following:
\begin{enumerate}
    \item with (p=0.1) return original nonaugmented image
    \item with (p=0.5) rotate image random angle in [-45, 45]
    \item with (p=0.5) rotate image on one of 90, 180, 270 degrees
    \item with (p=0.5) do random crop. We limit crop to retain all barcodes on the image entirely and ensure that aspect ratio is changed no more than 70\% compared to the original image
    \item with (p=0.7) do additional image augmentation. For this purpose we used "less important" augmenters from heavy augmentation example from \textit{imgaug} library \cite{imgaug}.
\end{enumerate}

\section{Experimental results}
\subsection{Datasets}
The network performance was evaluated on 2 common benchmarks for barcode detection - namely WWU Muenster Barcode Database (Muenster) and ArTe-Lab Medium Barcode Dataset (Artelab). Datasets contain 595 and 365 images with ground truth detection masks respectively, resolution for all images is 640x480. All images in Artelab dataset contain exactly one EAN13 barcode, while in Muenster there may be several barcodes on the image.

\begin{figure*}
    \centering
    \begin{subfigure}[b]{0.125\linewidth}
        \includegraphics[width=1.0\textwidth]{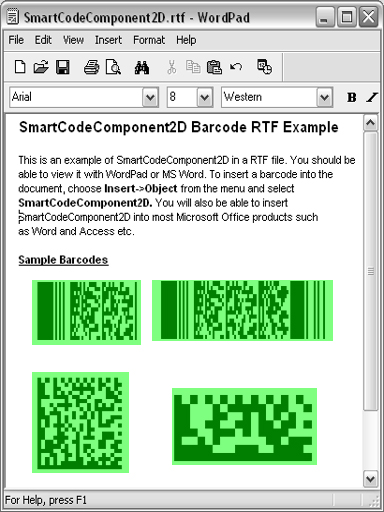}
    \end{subfigure}%
    \begin{subfigure}[b]{0.125\linewidth}
        \includegraphics[width=1.0\textwidth]{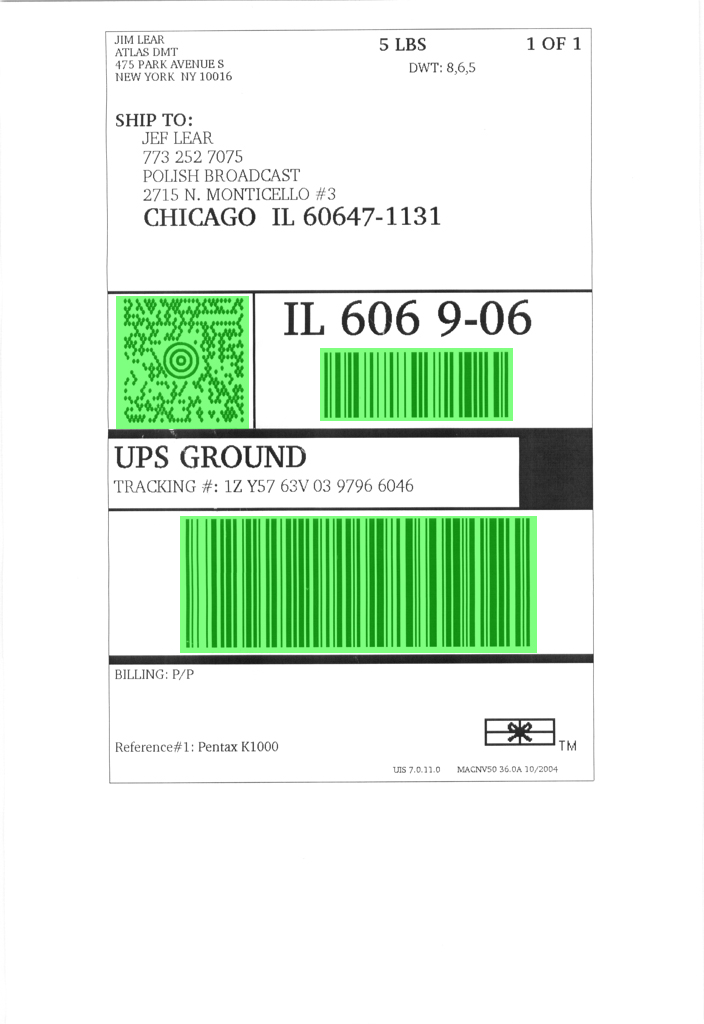}
    \end{subfigure}%
    \begin{subfigure}[b]{0.125\linewidth}
        \includegraphics[width=1.0\textwidth]{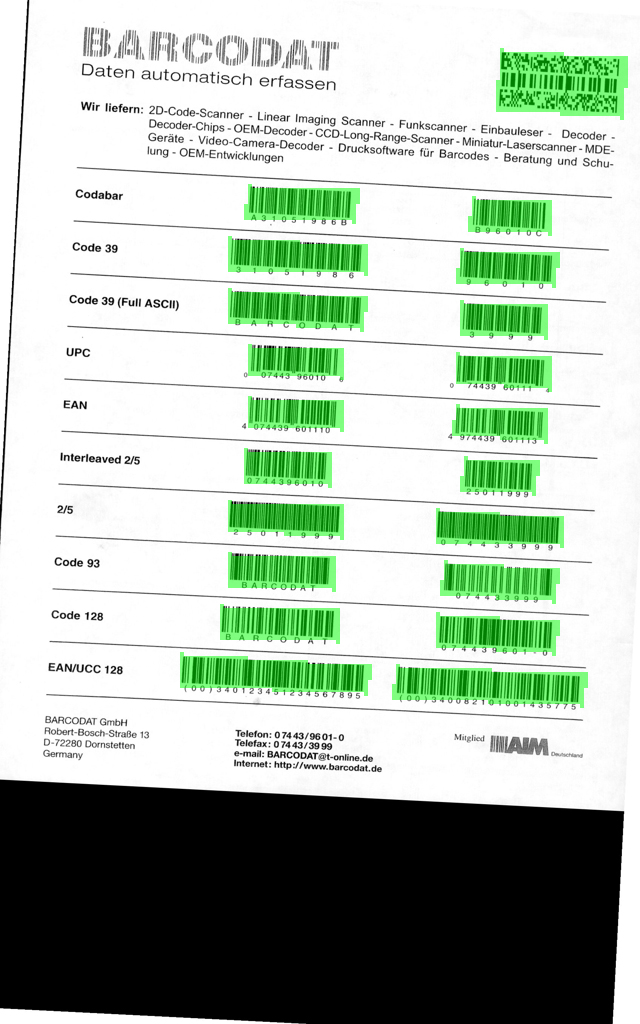}
    \end{subfigure}%
    \begin{subfigure}[b]{0.125\linewidth}
        \includegraphics[width=1.0\textwidth]{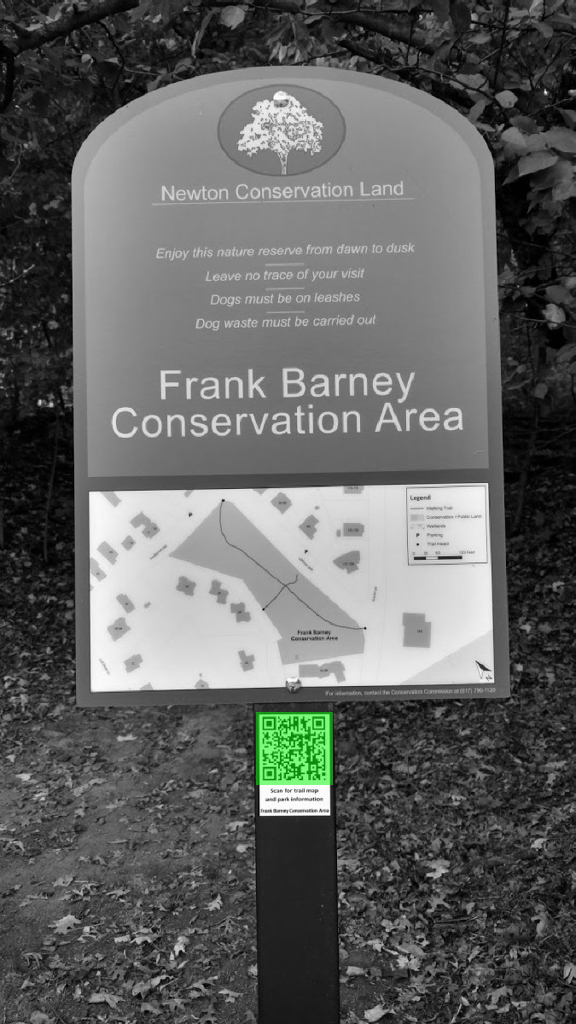}
    \end{subfigure}%
    \begin{subfigure}[b]{0.125\linewidth}
        \includegraphics[width=1.0\textwidth]{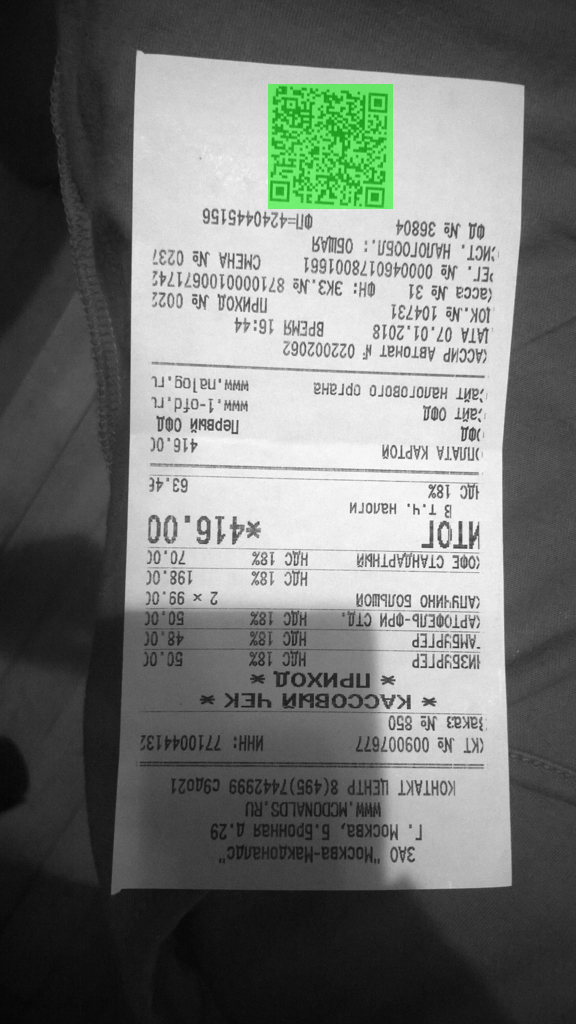}
    \end{subfigure}%
    \begin{subfigure}[b]{0.125\linewidth}
        \includegraphics[width=1.0\textwidth]{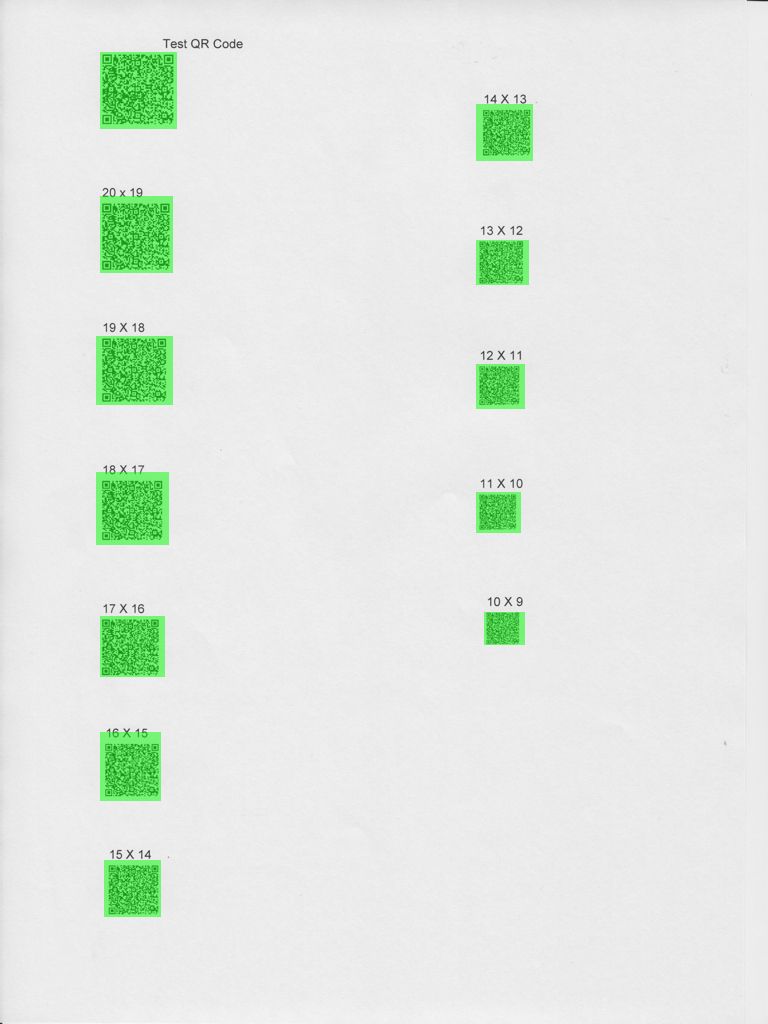}
    \end{subfigure}%
    \begin{subfigure}[b]{0.125\linewidth}
        \includegraphics[width=1.0\textwidth]{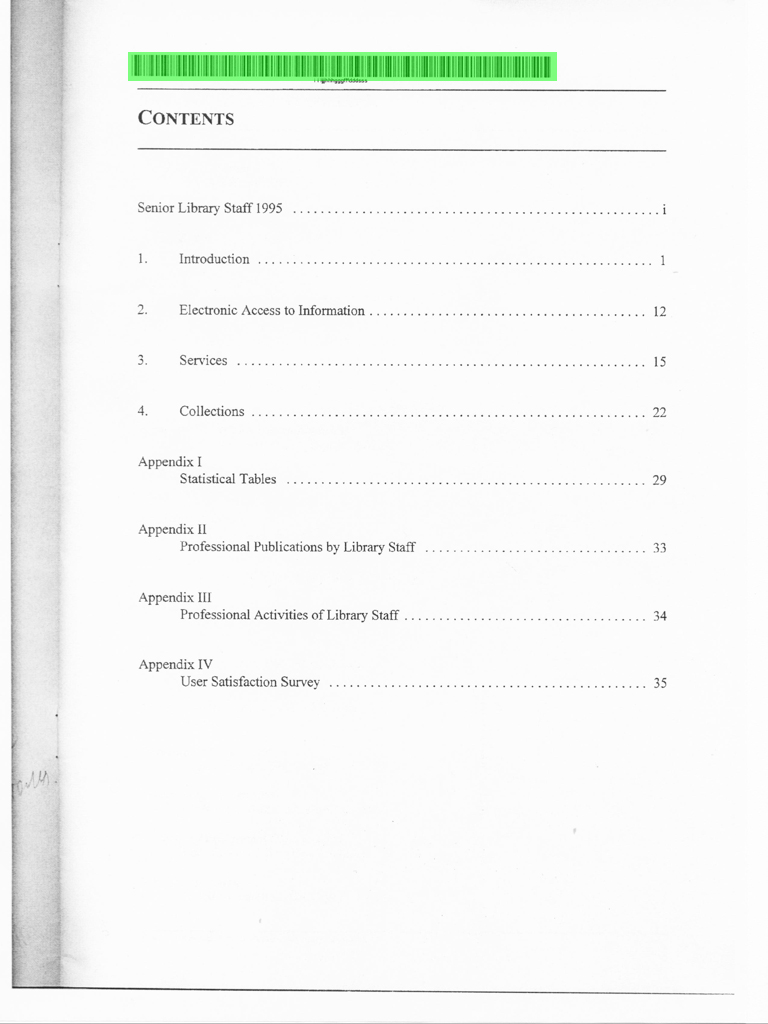}
    \end{subfigure}%
    \begin{subfigure}[b]{0.125\linewidth}
        \includegraphics[width=1.0\textwidth]{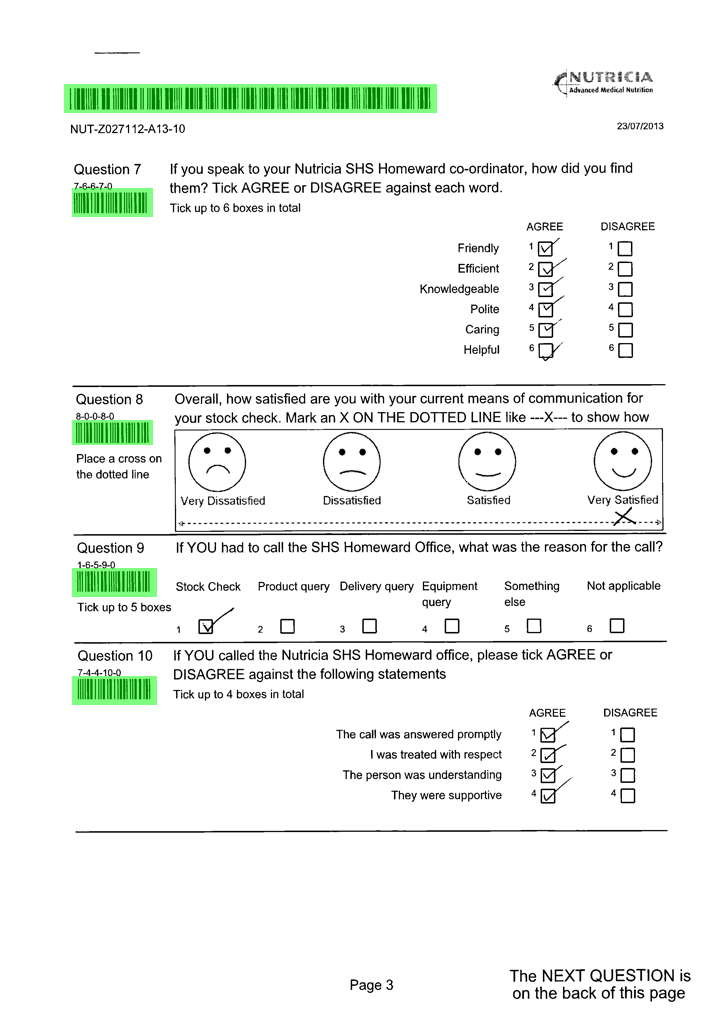}
    \end{subfigure}%
    \caption{Detection examples from test set}
    \label{fig:model_success}
\end{figure*}

\begin{figure*}
    \centering
    \begin{subfigure}[b]{0.2\linewidth}
        \includegraphics[width=0.99\textwidth]{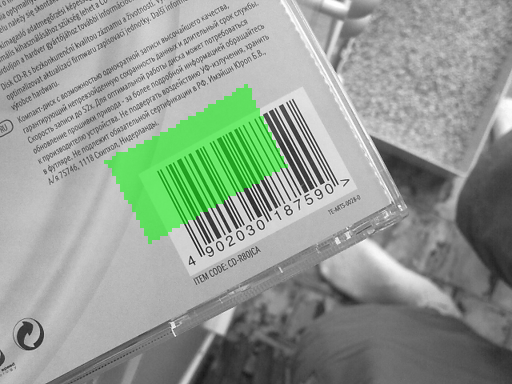}
        \label{fig:gull2}
    \end{subfigure}%
    \begin{subfigure}[b]{0.2\linewidth}
        \includegraphics[width=0.99\textwidth]{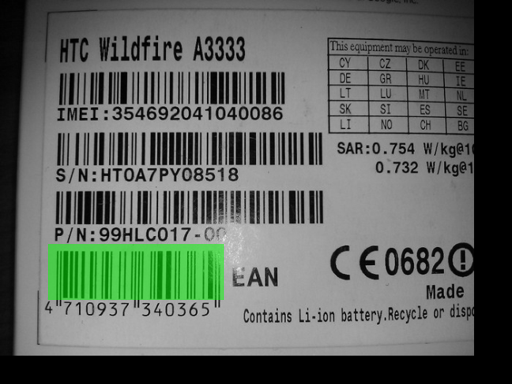}
        \label{fig:gull2}
    \end{subfigure}%
    \begin{subfigure}[b]{0.2\linewidth}
        \includegraphics[width=0.99\textwidth]{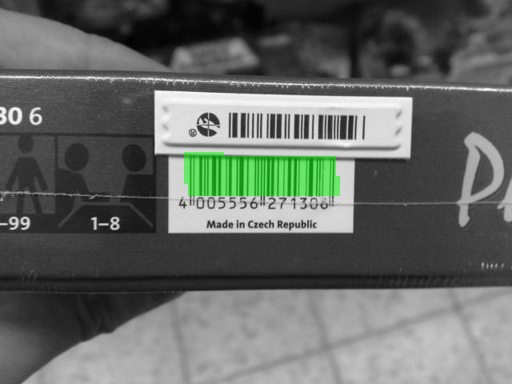}
        \label{fig:gull2}
    \end{subfigure}%
    \begin{subfigure}[b]{0.2\linewidth}
        \includegraphics[width=0.99\textwidth]{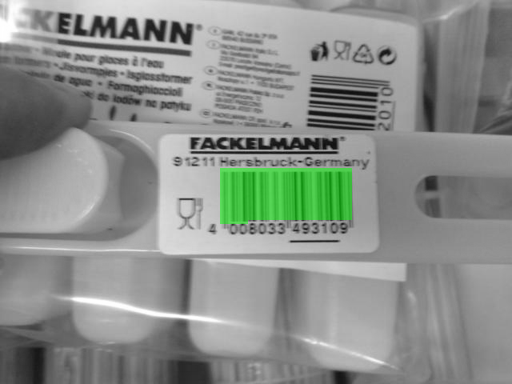}
        \label{fig:gull2}
    \end{subfigure}%
    \begin{subfigure}[b]{0.2\linewidth}
        \includegraphics[width=0.99\textwidth]{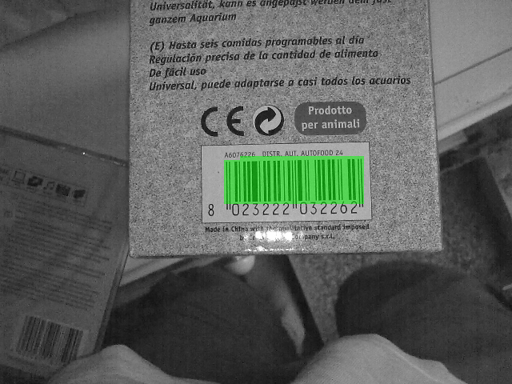}
        \label{fig:gull2}
    \end{subfigure}%
    \vskip -0.15in
    \begin{subfigure}[b]{0.2\linewidth}
        \includegraphics[width=0.99\textwidth]{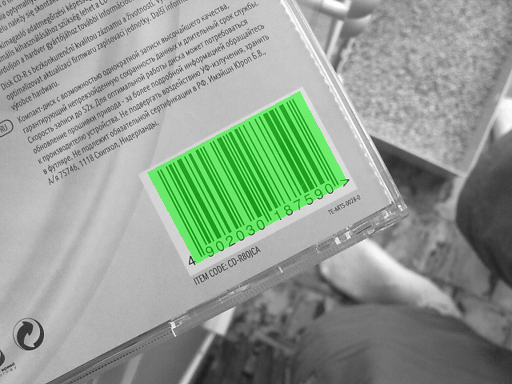}
        \label{fig:gull2}
    \end{subfigure}%
    \begin{subfigure}[b]{0.2\linewidth}
        \includegraphics[width=0.99\textwidth]{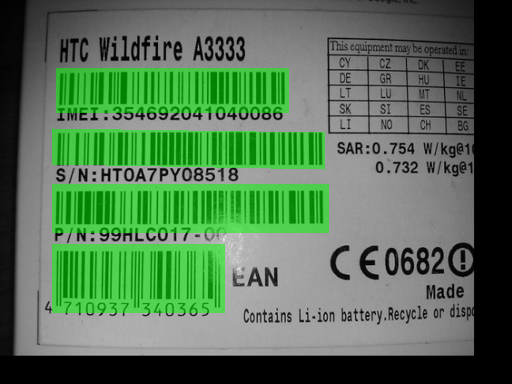}
        \label{fig:gull2}
    \end{subfigure}%
    \begin{subfigure}[b]{0.2\linewidth}
        \includegraphics[width=0.99\textwidth]{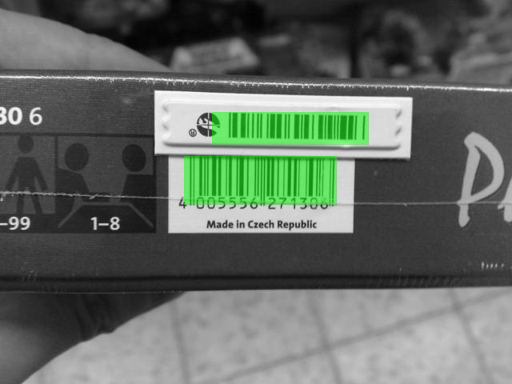}
        \label{fig:gull2}
    \end{subfigure}%
    \begin{subfigure}[b]{0.2\linewidth}
        \includegraphics[width=0.99\textwidth]{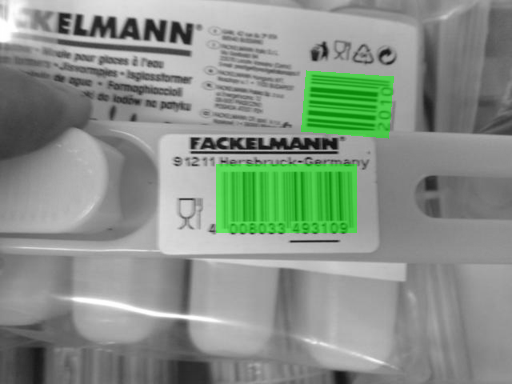}
        \label{fig:gull2}
    \end{subfigure}%
    \begin{subfigure}[b]{0.2\linewidth}
        \includegraphics[width=0.99\textwidth]{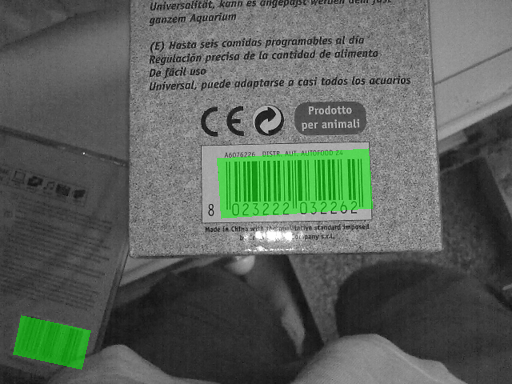}
        \label{fig:gull2}
    \end{subfigure}%
    \
    \caption{Detection examples: markup issues on Artelab and Muenster dataset. Markup on the top, detections results below}
    \label{fig:model_errors}
\end{figure*}

For training we used our own dataset with both 1D barcodes (Code128, Patch, EAN8, Code93, UCC128, EAN13, Industrial25, Code32, FullASCIICode, UPCE, MATRIX25, Code39, IATA25, UPCA, CODABAR, Interleaved25) and 2D barcodes (QRCode, Aztec, MaxiCode, DataMatrix, PDF417), being 16 different types for 1D and 5 types for 2D Barcodes, 21 type in total. Training dataset contains both photos and document scans. Example images from our dataset can be found in Fig.~\ref*{fig:model_success}. Dataset consist of 17k images in total, 10\% of it was used for validation.

\subsection{Training procedure}
We trained our model with batch size 8 for 70 epochs with learning rate 0.001 followed by additional 70 epochs with learning rate 0.0001

While training we resized all images to have maximal side at most 1024 maintaining aspect ratio and make both sides divisible by 64. We pick and augment/preprocess 3000 images from the dataset, then group them into batches by image size, and do this process repeatedly until the end of training. After that we pick next 3000 images and do the same until the end of the dataset. After we reach the end of the dataset, we shuffle image order and repeat the process.

We trained three models: \textit{Ours-Detection (all types)} (without classification on entire dataset), \textit{Ours-Detection+Classification (all types)} (with classification on entire dataset), \textit{Ours-Detection (EAN13 only)} (without classification on EAN13 subset of 1000 images).

\subsection{Evaluation metrics}
We follow common evaluation scheme from \cite{Cresot2015}. Having binary masks $G$ for ground truth and $F$ for found detection results the Jaccard index between them is defined as $$J(G,F) = \frac{|G \cap F|}{|G \cup F|}$$ Another common name for Jaccard index is "intersection over union" or IoU which follows from definition. The overall detection rate for a given IoU threshold $T$ is defined as a fraction of images in the dataset where IoU is greater than that threshold $$D_{T} = \frac{\sum_{i \in S} I(J(G, F) \geq T))}{|S|}$$ where $S$ is set of images in the dataset and $I$ is indicator function. 

However, one image may contain several barcodes and if one of them is very big and another is very small $D_T$ will indicate error only on very high threshold, so we find it reasonable to evaluate detection performance with additional metrics which will average results not by images but by ground truth barcode objects on them.

For this purpose we use recall $R_T$, defined as number of successfully detected objects divided by total number of objects in the dataset $$R_T = \frac{\sum_{i \in S}\sum_{G \in SG_i} I(J(G, F(G)) \geq T))}{\sum_{i \in S}|SG_i|}$$ where $SG_i$ is set of objects on ground truth on image $i$ and $F(G)$ is found box with highest Jaccard index with box $G$. The paired metric for recall is precision, defined as the number of successfully detected objects divided by total number of detections $$P_T = \frac{\sum_{i \in S}\sum_{G \in SG_i} I(J(G, F(G)) \geq T))}{\sum_{i \in S}|SF_i|}$$ where $SF_i$ is set of all detections made per image $i$.

We found connected components for ground truth binary masks and treat them as ground truth objects.

We emphasize that all the metrics above are computed for the detected object \textit{regardless its actual type}. To evaluate classification of the detected objects by type we use simple accuracy metric (number of correctly guessed objects / number of correctly detected objects). So if we find Barcode-PDF417 as Barcode-QRCode precision and recall will not be affected, but the classification accuracy will be.

\subsection{Quantitative results}

\begin{table*}[t]
\caption{Result comparation on different datasets.}
\label{table_artelab_muenster}
\begin{center}
\begin{small}
\begin{sc}
\begin{tabular}{l|cc|cc}
\toprule
  & \multicolumn{2}{c|}{Muenster} & \multicolumn{2}{c}{Artelab}\\
 & Acc $J_{avg}$ & Detection rate $D_{0.5}$ & Acc $J_{avg}$ & Detection rate $D_{0.5}$ \\
\midrule
Cresot2015 & 0.799 & 0.963 & 0.763 & 0.893 \\
Cresot2016 & - & 0.982 & - & \textbf{0.989} \\
Yolo2017 & 0.873 & \textbf{0.991} & 0.816 & 0.926 \\
Namane2017 & \textbf{0.882} & 0.966 & \textbf{0.860} & 0.930 \\
\midrule
Ours-Detection (all types) & 0.842 & 0.980 & 0.819 & \textbf{0.989}  \\
Ours-Detection (EAN13 only) & 0.762 & 0.987 & 0.790 & \textbf{0.995} \\
\bottomrule
\end{tabular}
\end{sc}
\end{small}
\end{center}
\end{table*}

\begin{figure}
    \centering
    \includegraphics[width=0.9\linewidth]{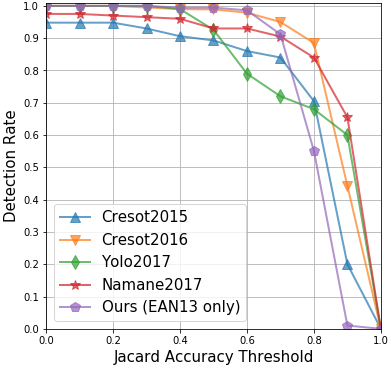}
    \caption{Detection rate for different Jaccard index thresholds}
    \label{fig:detection_rate_artelab}
\end{figure}

\begin{figure}
    \centering
    \includegraphics[width=0.8\linewidth]{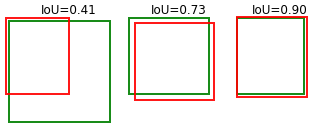}
    \caption{Example of Jaccard index computation for various bounding boxes}
    \label{fig:iou_examples}
\end{figure}

We compare our results with Cresot2015 \cite{Cresot2015}, Cresot2016 \cite{Cresot2016}, Namane2017 \cite{Namane2017}, Yolo2017 \cite{YoloBarcode} on Artelab and Muenster datasets (\autoref*{table_artelab_muenster}). 

The proposed method is trained on our own dataset, however all other works which we compared with were trained on different datasets. As for the full reproducibility of other authors works on our dataset we have to follow the exactly same training protocol (including initialization and augmentations) to not underestimate the results we decided to rely on the numbers reported in works of other authors.


We outperformed all previous works in terms of detection rate on Artelab dataset with the model trained only on EAN13 subset of our dataset. According to the tables, detection rate of our model trained on entire dataset with all barcode types is slightly worse than model trained on EAN13 subset. The reason for this is not poor generalization but markup errors or capturing more barcodes than in the markup (i. e. non-EAN barcodes), see Fig.~\ref*{fig:model_errors}. 

As it can be seen in Fig.~\ref*{fig:detection_rate_artelab} our model has a rapid decrease in detection rate for higher Jaccard thresholds. Aside from markup errors, the main reason for that is overestimation of barcode borders in detection, which is caused by prioritizing high recall in training, it makes high impact for higher Jaccard thresholds as Jaccard index is known to be very sensitive to almost exact match (Fig.~\ref*{fig:iou_examples}).

On \autoref*{table_precision_recall} we show comparison of our models by precision and recall. Our models achieve close to an absolute recall, meaning that almost all barcodes are detected. On the other hand precision is also relatively high.

\begin{table*}[t]
\caption{Precision and recall of our approach on different datasets, Jaccard index threshold set to 0.5.}
\label{table_precision_recall}
\begin{center}
\begin{small}
\begin{sc}
\begin{tabular}{l|cc|cc|cc|}
\toprule
 & \multicolumn{2}{c|}{Muenster} & \multicolumn{2}{c|}{Artelab} & \multicolumn{2}{c|}{Test Multiclass} \\
 & Precision & Recall & Precision & Recall & Precision & Recall \\
\midrule
Ours-detection (all types) & 0.777 & 0.990 & 0.814 & 0.995 & 0.940 & 0.991 \\
Ours-detection+classification (all types) & \textbf{0.805} & 0.987 & \textbf{0.854} & 0.995 & \textbf{0.943} & \textbf{0.994} \\
Ours-detection (EAN13 only) & 0.759 & \textbf{1.000} & 0.839 & \textbf{0.997} & - & - \\
\bottomrule
\end{tabular}
\end{sc}
\end{small}
\end{center}
\end{table*}

\subsection{Execution time}

\begin{table}[t]
\caption{Inference time comparison}
\label{table_time}
\begin{center}
\begin{small}
\begin{sc}
\begin{tabular}{lcc}
\toprule
 & Execution time (ms) & Resolution \\
\midrule
S\"{o}r\"{o}s \cite{Soros2013} & 73 & 960x723 \\
Cresot16 \cite{Cresot2016} & 40 & 640x480 \\
Yolo17 \cite{YoloBarcode} & 13.6 & 416x416 \\
Namane17 \cite{Namane2017} & 21 & 640x480 \\
Ours (GPU) & 3.8 & 512x512 \\
Ours (CPU) & 44 & 512x512 \\
\bottomrule
\end{tabular}
\end{sc}
\end{small}
\end{center}
\end{table}

For our network we measure time on 512x512 resolution which is enough for most of applications. We do not include postprocessing time as it is negligible compared to forward network run.

The developed network performs at real-time speed and is 3.5 times faster than YOLO with darknet \cite{YoloBarcode} on higher resolution on the same GTX 1080 GPU. In the \autoref*{table_time} we compare inference times of our model with other approaches. We also provide CPU inference time (for Intel Core i5, 3.20GHz) of our model showing that it is nearly the same as reported in Cresot2016, where authors used their approach in the real-time smartphone application. It is important since not all of the devices have GPU yet.

\subsection{Classification results}

Among correctly detected objects we measured classification accuracy and achieved 60\% accuracy on test set.

Moreover, classification subtask does not damage detection results. As shown in \autoref*{table_precision_recall} the results with classification are even slightly better, meaning that detection and classification tasks can mutually benefit from each other.

\subsection{Capturing long narrow barcodes}

Additional advantage of our detector is that it is capable of finding objects of any arbitrary shape and does not assume that objects should be approximately squares as done by YOLO. Some examples are provided in Fig.~\ref*{fig:model_success}.

\section{Conclusion}

We have introduced new barcode detector which can achieve comparable or better performance on public benchmarks and is much faster than other methods. Moreover, our model is universal barcode detector which is capable to detect both 1D and 2D barcodes of many different types. The model is very light with less than 33000 weights which can be considered very compact and suitable for mobile devices.

Despite being shallow (i.e. very simple, we didn't use any SOTA techniques for semantic segmentation) our model shows that semantic segmentation may be used for object detection efficiently. It also provides natural way to detect objects of arbitrary shape (e.g. very long but narrow).

Future work may include using more advanced approaches in semantic segmentation to develop better network architecture and increase performance.


\bibliographystyle{IEEEtran}
\bibliography{main}
\end{document}